# LA-UR-25-31130

**Approved for public release; distribution is unlimited.**

| | |
|---|---|
| **Title:** | Assessing the Applicability of Natural Language Processing to Traditional Social Science Methodology: A Case Study in Identifying Strategic Signaling Patterns in Presidential Directives |
| **Author(s):** | LeMay, Chase Joseph<br>Lane, Adam Martin<br>Seales, Johnny<br>Winstead, Maria<br>Baty, Samuel R. |
| **Intended for:** | Report |
| **Issued:** | 2025-11-12 (Draft) |

![Los Alamos National Laboratory logo]

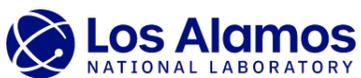
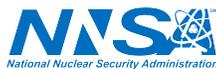
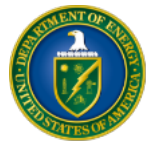

Assessing the Applicability of Natural Language Processing to Traditional Social Science Methodology: A Case Study in Identifying Strategic Signaling Patterns in Presidential Directives

By Chase LeMay, Adam Lane, Johnny Seales, Maria Winstead, and Samuel Baty

## Table of Contents

## Abstract

Our research investigates how Natural Language Processing (NLP) can be used to extract main topics from a larger corpus of written data, as applied to the case of identifying signaling themes in Presidential Directives (PDs) from the Reagan through Clinton administrations. Analysts and NLP both identified relevant documents, demonstrating the potential utility of NLPs in research involving large written corpuses. However, we also identified discrepancies between NLP and human-labeled results that indicate a need for more research to assess the validity of NLP in this use case. The research was conducted in 2023, and the rapidly evolving landscape of AIML means existing tools have improved and new tools have been developed; this research displays the inherent capabilities of a potentially dated AI tool in emerging social science applications.

## Introduction

States have interests they are willing to defend and lines they will not allow to be crossed without retaliation. It is often in the state's interest to communicate these boundaries and potential responses so that adversaries do not unwittingly create a conflict that would incur costs for which they were not prepared. This is especially important in the context of nuclear weapons, where the cost could be catastrophic; a state may deter adversaries by stating that any established boundaries would be defended to the fullest extent, and whether that is true or not these claims must be backed up by observable measures to indicate the state's willingness and capability to follow through. These observable measures are referred to as strategic signaling. To better understand how strategic signaling has been used by different US presidential administrations, we used human-assisted natural language processing in coordination with analysis conducted by researchers to explore Presidential Directives (PDs) to see how these documents were used to direct strategic signaling and to look for trends across administrations.

The first aim of our research was to determine if natural language processing (NLP) could be used as a tool by researchers to quickly identify topics in written data, specifically in the context of national security policy documents. By creating a case study around the language used in PDs, we were able to compare the themes found by NLP and its categorization of documents against the themes and categorizations we as analysts had made. Specifically, we wanted to see if the data collection phase of social science research could be expedited through the application of latent Dirichlet allocation (LDA) topic modeling, a specific form of NLP which extracts latent topics from within a written dataset.

The second aim of our research was to determine if there were trends in the focus of nuclear strategic signaling by different presidential administrations, spanning from Reagan to Clinton, looking specifically at the signaling directed at the Soviet Union/Russia. We chose to analyze PDs because these documents are one of the ways in which the president directs traditional signaling activity, such as diplomatic engagements, military planning and budgeting. Therefore, although PDs are intended primarily for an internal audience, we considered the language used in these documents as indicative of an administration's overall strategy. Our timeframe starts with the Reagan administration, as we chose to examine the decade before and after the end of the Cold War and dissolution of the Soviet Union. Within these documents, we identified several relevant themes and theorized that if they could be analyzed across time, correlations could be made to strategic goals, and successes and lessons could be drawn for contemporary signaling situations.

In the following pages, we review some of the existing research on strategic signaling as well as NLP. We then explain the methodology we used for our research before detailing our findings.

## Selected Literature Review: Signaling and Natural Language Processing

Regarding natural language processing, the value of topic modeling (TM), which uses machine learning programs to organize words or phrases into themed clusters for use in social science research, was proposed by Valdez, et al in 2018.[1] In *Topic Modeling: Latent Semantic Analysis for the Social Sciences*, they argue that TM, which encompasses various models for uncovering themes in large collections of text documents, known as a corpus, is particularly valuable to the work of social scientists but has been underutilized.[2] Among the benefits of TM over traditional content analysis is a reduction in researcher subjectivity. TM is also capable of word disambiguation, making distinctions between words that are spelled the same but have different meanings; a capability not found in traditional word-count methods. Valdez, et al focus on one TM model, latent semantic analysis (LSA), which they believe is especially well suited to social science research; using LSA, they analyzed transcripts of 2016 presidential debates between then presidential candidates Hillary Clinton and Donald Trump.[3] As the authors expected, they found that the themes generated by LSA in the analysis of the debate transcripts mirrored the most commonly searched/Googled policy-related terms prior to the election, and thus argued that LSA's ability to find themes in the transcripts, which matched salient political issues, supported the validity of the model.[4]

Another TM model, latent Dirichlet allocation (LDA), was explored by Blei, et al in 2003.[5] LDA is similar to LSA in that it can be used for classification, summarization, similarity, and relevance judgements, and is capable of differentiating between words which are spelled the same but have different meanings.[6] Blei, et al argue that LDA offers an advantage over simple Dirichlet-multinomial cluster modeling in that unlike a classical clustering model which can only associate a document with a single topic, LDA can associate a document with multiple topics. LSA can also recognize that a document can contain multiple topics but is not as capable as LDA in addressing topics not seen in training documents and is more prone to problems of overfitting than LDA.

There are other traditional NLP techniques applicable to social science research, and cutting-edge technology may transform this analysis as new capabilities emerge. Techniques such as Named Entity Recognition, which can identify entities (people, places, dates, etc.) in a corpus, and Coreference, which identifies when entities that have been previously mentioned are being referred to (e.g. going "there" refers to going to "Paris"), could completely change how written data is processed. Additionally, fine-tuning a large language model (LLM), a type of AI which is trained with huge sets of data to interpret written data and language and can generate text and complete other tasks, could improve performance of NLP programs by providing additional training materials on a specific topic the LLM should be familiarized with.[7] For example, an LLM tasked with identifying signaling patterns in a new corpus of documents would benefit from being "fine-tuned" with documents/written data commonly affiliated with signaling.[8]

The role of signaling in foreign policy has been the topic of much analysis. In his paper "Signaling Foreign Policy Interests", James Fearon looks at two ways states can signal their willingness to follow through with a threat.[9] Fearon explains that although a state may be capable of, and willing to, follow through with a threat of force, it would generally prefer not to use force. The challenge for a state then, is to credibly signal to an adversary the state's willingness to follow through on threats in defense of vital interests, without incurring the costs associated with making good on these threats. This can be done in two ways according to Fearon; tying hands or sinking costs. Tying hands involves making public statements that would then cause a leader to incur political costs from his audience if he were to back down from a threat made publicly during a crisis. Sinking costs involves taking financially costly actions, such as mobilizations or arming measures, to signal a willingness to use force to defend a state's interest. While Fearon explains that states will often use a combination of these methods, he finds that leaders are more likely to rely on tying hands during a crisis scenario and more likely to use sinking costs for longer-term grand strategy signaling.

For the purpose of our research, we built on John Gower's work in "Nuclear Signaling Between NATO and Russia," which outlined signaling as "send[ing] signals about [a nuclear state's] capability and intent, aimed at three important though largely unconnected audiences: domestic populations, allies, and current and potential adversaries" that "follow different timeframes: from long-term strategic signaling to short-term reactive activity and crisis signaling."[10] The importance of signaling in the specific context of nuclear deterrence was explored by Lindsay and Gartzke in their edited volume *Cross Domain Deterrence, from Practice to Theory* in 2019.[11] Although this work focuses on "cross-domain deterrence", a concept that arose during the George W. Bush Administration to address the complex threat environment of the 21st century, it includes concepts that are useful when considering nuclear signaling more broadly. For example, the authors state that "The central paradox in both theory and practice of deterrence was how to credibly threaten to use weapons for coercion (i.e., deterrence or compellence) that were too costly to use in war."[12] Chapter 10 of this book was written by Green and Long and explores the balance necessary to signal nuclear capabilities without giving away secrets which would allow US adversaries to develop technologies to counter US capabilities.[13] The authors conclude that this was done successfully in the late Cold War period through a combination of military demonstrations, selective leaks, counter-intelligence operations, and public statements. For example, the authors state that the Carter-era Presidential Directive 59 was intentionally leaked to the New York Times so that the Soviet Union would be aware of the US missile targeting of Soviet political leadership priorities.[14] Through means such as these, the US was able to demonstrate to the Soviets vulnerabilities in their systems, such as their early warning, C3, and SSBNs, without revealing the specific technical means through which US capabilities worked.

## Methodology

NLP was recognized as a potential tool to reduce the time necessary to identify relevant documents in a large dataset. Our goal was not to prove that NLP could replace analysts in reading and processing written data, but rather to evaluate whether NLP, particularly an LDA topic modeling program using Human-in-the-Loop (HITL) methods, could assist analysts in identifying relevant documents in large datasets to facilitate further research. Therefore, our focus was to identify relevant topics in documents and compare analyst generated results with those of the topic modeling program.

---

[10] Gower, John. *Nuclear Signalling Between NATO and Russia*. European Leadership Network, 2018. *JSTOR*, http://www.jstor.org/stable/resrep22136. Accessed 6 June 2024.
[11] Lindsay, Jon R, and Erik Gartzke, *Cross Domain Deterrence, from Practice to Theory* (Oxford University Press, 2019), 1. Accessed 28 September 2023.
[12] Lindsay, Jon R, and Erik Gartzke, *Cross Domain Deterrence, from Practice to Theory* (Oxford University Press, 2019), 1. Accessed 28 September 2023.
[13] Green, Brendan Rittenhouse and Austin G. Long, *Signaling with Secrets: Evidence on Soviet Perceptions and Counterforce Developments in the Late Cold War*, (Oxford University Press, 2019). Accessed 28 September 2023.
[14] Green, Brendan Rittenhouse and Austin G. Long, *Signaling with Secrets: Evidence on Soviet Perceptions and Counterforce Developments in the Late Cold War*, (Oxford University Press, 2019). Accessed 28 September 2023.

### Gathering Data

We began by searching for all PDs from the Reagan administration onward. Unfortunately, our review of available PDs revealed a significant drop-off in publicly available PDs after the Clinton administration. Most PDs following Clinton remain classified or are awaiting declassification review. Subsequently, we chose to restrict our analysis to the publicly available PDs from the Reagan through Clinton administrations. All results are based exclusively on unclassified PDs currently available via presidential libraries with a data cutoff date of June 2023.

### Choosing Relevant NLP Tools

To apply NLP to our research model, our team worked with a data scientist to map the functions of the program against what we were trying to identify in our data. We chose an LDA topic modeling program to identify topics that were consistent across the PDs and isolate ones we believed were relevant and used an HITL process to refine the results. We anticipated NLP would not be able to replace analysts in its ability to understand intricacies in written data, particularly when inconsistent phrasing or abbreviations interrupt the program's ability to cluster topics, but that it should be capable of identifying relevant themes and topics within the documents. By using HITL, analysts could adjust the model to fit the framework of our project.

### Category Creation

The topic search process for the NLP was designed to be similar to the analyst process. Using an understanding of signaling as explored by Gower, Lindsay, and Gartzke, we created seven categories of signaling we sought to identify in the documents: Threat of Force, Movement of Forces/Materials, Maintenance, Funding, Programs, Reductions/Arms Control, and Monitoring/Verification. These categories reflect the range of US nuclear enterprise priorities in the selected timeframe, and we discussed associated language which would help us sort documents into those categories. A breakdown of analyst instructions and a detailed list of the research process are available in Annexes 1 and 2, respectively.

Once the categories were created, the NLP and analyst review processes were conducted concurrently.

### Analyst Categorization/Binning

In the first stage, the analysts quickly reviewed each PD to determine if it related to nuclear issues and if it related to signaling directed toward the Soviet Union/Russia. Instructions to review documents were as follows: judge from the title, early sections, a quick eyes-only document scan, and the search function of the digital file (when available) to determine if the document contained any of the relevant categories. This process was intended to expedite the processing of large amounts of written data. Analysts used a layered approach in which each analyst reviewed the documents and evaluations of the other analysts in a rotating order.

Analysts were each initially responsible for a portion of the documents to indicate Yes/No/Maybe if the document generally applied to the nuclear enterprise. If it received either a Yes or Maybe, it was pushed forward to the next phase of our review process where it was reviewed by another analyst who confirmed/challenged the decisions of the first analyst. The second analyst also identified one dominant category in the text and prepared a brief summary. Those documents were then reviewed by a third analyst who performed the same review with a summary and identification of all relevant categories. This layered method was intended to allow analysts to challenge the judgment of their peers and remove irrelevant documents from the data pool while reducing time spent reading documents. Our team determined that this method would be sufficient across most documents to establish topics represented within the PDs. In the third phase, analysts could mark as many categories per PD as they deemed appropriate.

*NLP Topic Generation*

To apply NLP to our research, we needed to select the number of topics that the program would identify in the corpus. We determined an output of 40 topics was sufficient to mitigate false groupings of related topics while also alleviating concerns that some topics would be unnecessarily divided. Requesting additional topics may have provided increasingly granular divisions of information which could have resulted in more work for the analysts to individually assign, while requesting fewer topics may have unnaturally grouped information together under one topic and reduced the impact of using the tool as analysts would have needed to manually separate the relevant data. This was the first human element to be introduced into the NLP process and ensured that the data produced in the next steps would be relevant to our project. Using HITL NLP, particularly in this scenario with 40 requested topics, allowed for analysts to quickly determine relevant categories without using an impractical amount of time to do so.

The topic modeling program identified the 40 requested topics, previewing the top ten words it associated with a topic with the option to list additional associated words for a detailed collection. The analysts were then responsible for reviewing the lists and determining which of the seven analyst identified categories they corresponded to. We applied the same categories to the NLP that the analysts had identified to maintain consistency. We determined, however, that the best way to show how NLP identified topics similarly to analysts would be to create a weighted system where each topic had a ranked categorization. Analysts discussed what language they considered relevant to the NLP generated topics to inform a hierarchical categorization of topics, deciding which category would be the primary, secondary, and tertiary for each topic. See the results of this process in the image provided on page 10.

For example, Topic 0 was identified with the following topic keywords: missile, base, strategic, missiles, force, cruise, peacekeeper, minuteman, deploy, and development. Analysts decided that this topic would be categorized as Programs, Movement of Forces/Materials, and Threat of Force in that order.

## *NLP Topic Categorization*

Analyst categorization of NLP generated topics constituted the most influential aspect of the HITL process, as it had a direct impact on how those topics would be reflected in our data outputs. The hierarchical structure of the categorization of topics was integrated into our research approach to reflect the ability of words/phrases to have multiple meanings. In the earlier described analyst-only approach to processing documents, analysts might read certain phrases or topics as being applicable to multiple categories, which was reflected by analysts' ability to choose as few or as many categories as necessary. In an attempt to mirror that capability in the HITL NLP approach, each relevant topic keyword produced by the NLP was then, by a human analyst (the HITL part of HITL NLP), assigned three categories where the order influenced that topic's influence on the weighted category frequency output i.e. the category with the highest connection to the topic was ranked first, the runner-up was ranked second, and the weakest relevant connection was ranked third. All documents that analysts did not deem relevant were not ranked, receiving an "Other" designation for all three rankings. The topics and their rankings are shown on the page below.

## *NLP Document Identification Instructions*

Our goal was for the program to provide the following for each document: the frequency of each topic, the three most prevalent categories (if the document was relevant), and a simple indication of whether the document should be included in a focused data set for analysts to read. Recognizing the wide scope of relevant topics, we had to decide which terms we wanted the program to focus on. A significant number of PDs focused on foreign policy priorities and methods to promote U.S. interests abroad, so we realized we would need to cast a wide net to capture as many relevant documents as possible in the NLP's "analyze_document" category. At the same time however, many documents related to foreign policy were not related to the key nuclear aspect of our research, so we had to find a way to try to only capture documents that met the conditions the analysts had been searching for.

Our solution was to have the program scan for the word "nuclear" in conjunction with the topic breakdown to determine if a document contained elements pertaining to our research. Based on the scope of our research, we discussed if adding additional words to the scan that related to our focus (e.g, Soviet, strategic, missile, Russia) could help narrow the results, but we elected to only use the term "nuclear" as to not accidentally block relevant documents without additional terms. This turned out to be prudent as there were many documents the analysts determined to be relevant that were not flagged by the NLP that were missing the term "nuclear" from the text file. Had we added additional phrases, many documents that we identified as pertaining to our research would likely not have been flagged by the NLP. Additionally, there were many documents that the NLP flagged as being relevant that were not related to our topic that mentioned the term "nuclear" but were unrelated to our focus.

| Topic Number | Topic Keywords (Top 10) | Topic Label | 1 | 2 | 3 |
|---|---|---|---|---|---|
| 0 | missile base strategic missiles force cruise peacekeeper minuteman deploy development | deployment of strategic missiles | 5 | | |
| 1 | security state national follow policy house review number white agreement | action: policy review process | 7 | | |
| 2 | soviet reductions arm negotiations union side agreement nuclear table | arms control negotiations / reductions with Russia | 3 | 5 | 7 |
| 3 | deployment data nuclear weapons force level restrict authorization secret table | deployment of nuclear weapons | 2 | 6 | 7 |
| 4 | nstd policy supersede emergency preparedness civil emergencies nsep mobilization federal | preparedness for civil emergencies | 7 | | |
| 5 | secretary state director president national security unite staff defense control | staffing roles/assignment: policy participants | 7 | | |
| 6 | strategic force soviet nuclear program defense ally offensive union treaty | strategic defense against Russian nukes | 5 | 6 | |
| 7 | export transfer technology license policy arm control case coom commerce | Technology export control | 7 | | |
| 8 | africa south china military african marta chinese settlement government angola | Reduce soviet influence on China/Angola | 7 | | |
| 9 | military force lebanon libya iraq libyan israel gulf government iran | Middle east stability | 7 | | |
| 10 | soviet restraint union interim salt unite mutual take state tcbm | Mutual restraint during negotiations | 5 | 3 | 7 |
| 11 | make period clear elimination year ahon soviet would area pace | Arms control timescales | 5 | 3 | 7 |
| 12 | start heavy would aloms sdoms side mobile bombers nuclear arm | Start negotiations delivery systems | 5 | 3 | 2 |
| 13 | central america cuba government caribbean democratic nicaragua action plan secretary | Stability in central America and caribbean | 7 | | |
| 14 | operations peace peacekeeping force operation military cinpol support humanitarian police | Humanitarian and peace keeping operations | 7 | | |
| 15 | nuclear weapons stockpile plan production materials weapon part program security | nuclear weapons stockpile planning | 3 | 1 | 5 |
| 16 | soviet arm control union report test verification compliance noncompliance treaty | soviet arms control noncompliance | 4 | | |
| 17 | would make also provide appropriate seek could take base include | evaluating programs and opportunities (funding and projects) | 7 | | |
| 18 | security national policy shall group establish president review secretary interagency | Staff that establish and evaluate national security policies | 7 | | |
| 19 | economic state support interest continue political unite objectives policy security | continued economic support for US national security and policy objectives | 7 | 2 | |
| 20 | environmental state arctic unite official international management development policy sustainable | environmental protections of arctic region | 7 | | |
| 21 | supply energy market emergency drug disruption narcotics demand price production | price reduction of energy and drugs | 7 | | |
| 22 | unite state weapons nuclear control proliferation chemical destruction technology biological | nuclear, chemic and biological nonproliferation | 5 | 3 | |
| 23 | project crisis management system medussa nrccs white phase house president | crisis information management systems | 7 | | |
| 24 | treaty group negotiate state offensive strategic space defense proposal agreement | strategic space defense treaty | 3 | 5 | 7 |
| 25 | inspection team section site inspections treaty lead party provision osia | site inspections | 4 | 3 | 2 |
| 26 | state shall government intelligence department foreign agencies activities appropriate unite | defense against and collection of foreign intelligence | 7 | | |
| 27 | national infrastructure sector private critical shall information official federal coordinator | crtical national infrastructure | 7 | | |
| 28 | security information systems national telecommunications executive directive classify agent technical | security of telecommunications | 2 | 5 | |
| 29 | test nuclear program energy treaty verification stockpile treaties monitor cooperation | programmatic approval of nuclear testing | 5 | 7 | |
| 30 | port state vessels claim unite right security program navigation maritime | freedom of maritime navigation program | 7 | | |
| 31 | program trade economic japan food level private countries investment debt | bilateral economic relations with japan | 7 | | |
| 32 | summit meet economic white group house preparations president coordinate representative | White house preparations for economic summit | 7 | | |
| 33 | terrorism terrorist force task group citizens recommendation act terrorists lead | combatting terrorism | 7 | | |
| 34 | soviet union relations relationship soviets pakistan ussr moscow east afghanistan | support for Afghanistan resistance against Russia | 7 | | |
| 35 | program defense plan provide support military national ensure capabilities capability | Strategic planning of military capabilities | 5 | 7 | |
| 36 | redact broadcast international information radio government programs audiences thesecretary marti | information security of private communications (encryption) | 7 | | |
| 37 | shall action president covert find security advisor national conduct intelligence | Conducting intel operations abroad | 7 | | |
| 38 | missiles limit ballistic missile warheads start deploy idoms number additional | Limits on deployed ballistic missiles | 3 | 1 | 5 |
| 39 | space launch commercial national unite civil government foreign state security | National security of civil and commercial space use | 7 | | |

Key

| Value | | Category |
|---|---|---|
| 7 | | Other |
| 6 | | Funding |
| 5 | | Programs |
| 4 | | Mon/Ver |
| 3 | | Arms Control |
| 2 | | Maintenance |
| 1 | | Movement |
| 0 | | ToF |

Combining the two decision factors, we decided the best metric for the NLP output to determine if a document should be considered relevant for the analysts was for the following conditions to be true: IF contains_nuclear=1 AND other_dominates=0 THEN analyze_document=1. We ran the program using these parameters and initial data showed a 2.97% numerical difference in documents identified as relevant; the NLP had identified 146 documents that met the conditions, in comparison to the 134 that analysts had identified.

However, there were several factors in how we applied NLP that impacted the outputs of the data. Firstly, while we intended for the NLP to categorize any document that met the conditions outlined above, there were multiple documents that met those conditions but were not categorized or were categorized but did not meet the conditions to be marked as analyze_document=1. This resulted in our NLP identification total being different from our NLP categorization total. While reviewing this phenomenon within the results, we realized that this was due to how the program categorized documents with a majority of language being identified within the Other category.

- In some cases, the Other category constituted the plurality of the document but not the majority (> .50) and was therefore eligible for categorization and analyze_document=1.
- In some cases, the Other category constituted the majority of the document but the language within the document should have reflected enough to be categorized by the program but wasn't.

*Errors and Interference*

During the identification process, the NLP method created two types of errors; Type 1 errors occur when a paper was identified by the NLP model as being relevant to one of the categories of interest but turned out not be relevant. Type 2 errors are the converse – instances when the model did not identify the document as being relevant, but human coding identified it as relevant. Both types of errors require correction (and can be damaging to an automated process), but in most topics, Type 2 errors are more problematic to correct. Type 2 errors create a situation in which data is excluded, sometimes without an additional review, creating an error that is often more persistent than a Type 1 error.

To use the topic modeling program, all uploaded files had to be searchable text documents rather than scanned image files. For many PDs, the files uploaded to the presidential libraries were able to be directly transferred into a text file format (.txt) for processing. However, if the search function could not be used on a PD and the base PDF version could not be scanned by the NLP, we used optical character recognition (OCR) which is a tool that identified characters on a scanned page and converted them into text.

There were multiple instances where markings on the document interfered with both the NLP model and analysts' abilities to read the document. These interferences typically came in the form of stamps onto the original document preventing visibility of words. While having a substantive impact on both analyst and NLP reading of documents, we did not consider PDs with redacted text as corrupted in this evaluation as those were intentionally covered as part of the release of the document. Examples are provided below with varying levels of data interference. While some of the interference impacted analysts' ability to read the document, it had a much greater impact on the NLP's ability to pull text from the documents. Analysts reviewed each document and determined that 53 of the 404 documents had markings, and approximately 13% of our documents required using OCR to obtain our data. Examples of interference can be seen below.

## Minimal and Low-Level Interference

The following pictures show examples of minimal and low-level corruption of the data, which did not significantly affect the data.

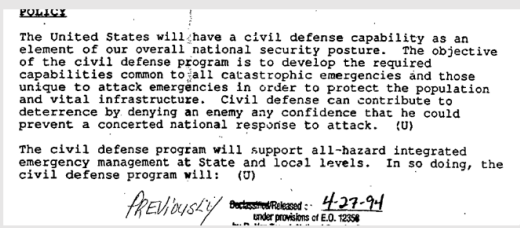

Minimal-impact interference

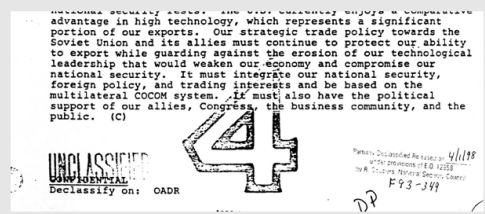

Low-impact interference

## Medium and High-Level Interference

The following pictures show examples of medium and high-level corruption of the data, which significantly affected the scanned data.

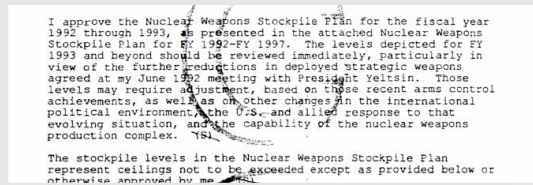

Medium-impact interference

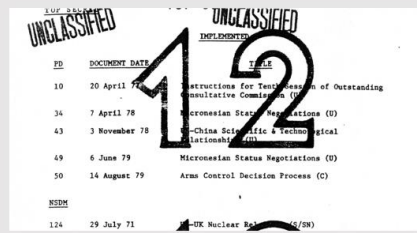

High-impact interference

### *Discrepancies*

There were 59 discrepancies between analyst evaluation and that of the NLP, accounting for 14.6% of all documents. There were 23 documents which the analysts identified as relevant which the NLP did not, and 36 documents the NLP identified as relevant that the analysts did not. In review of each document, we determined the following as the main reasons for discrepancies between analyst and NLP evaluations:

- The NLP identified documents that were not related to our project, a limitation induced by only including the term "nuclear" as indication of a relevant document

- In a review of the 36 discrepancies where NLP flagged documents we had not, we determined that in 28 out of 36 cases the NLP had flagged documents meeting the criteria but unrelated to our focus.
- Of the 36 documents identified by the NLP that analysts had not, 13 of the documents were sorted into categories by the NLP while the 23 remaining were categorized as Other.
  - All documents categorized as Other had contains_nuclear=1 and other_dominates=0, but nearly ⅔ of the documents did not have enough significant language to fit into the categories
- The analysts identified documents that were related to our topic through existing knowledge, where the documents did not meet both conditions outlined for analysis
  - In a review of the discrepancies where analysts flagged documents NLP had not, we determined that in 18 out of 23 cases analysts had flagged documents that were related to the focus but did not meet the criteria.
  - Of the 23 documents analysts identified but NLP had not, 10 of the documents had been sorted into categories by the NLP while the remaining 13 were categorized as Other.
- The NLP identified documents that analysts failed to identify or misidentified
  - These were the result of human error during our identification process, where a combined 13 documents were improperly identified as either false positives or false negatives by analysts.

After properly classifying the documents which we verified NLP had identified correctly, we created a Confusion Matrix (Page 14) to display the results of the discrepancies.

## Results

In a review of all the documents, the NLP had an accuracy of 88% (358/404) identifying PDs as relevant or irrelevant to our research and had a precision of 80% (118/146) in identifying documents that analysts should have read. These results indicate that even in an unrefined format with general instructions, LDA topic modeling provided analysts with valuable insights into the overall contents of the documents and could likely assist in the data gathering process of social science research using national security documents.

Proportionally, more Type 1 errors occurred than Type 2 – by over a factor of ~1.5. The authors hypothesize that some of these errors are caused by obstructions to the ability for the NLP to fully analyze the document (due to artifacts such as stamped markings on the pages). Still, significant model refinement is needed to also avoid Type 2 errors, as the model should be biased towards providing analysts with all possibly relevant information before NLPs can be used without robust human oversight of the data.

In situations where the NLP properly identified a document where analysts did not, the documents were counted here as True Positive or True Negative accordingly.

## Confusion Matrix

| | True Positive: | False Positive: |
|---|---|---|
| | **118** | **28** |
| | False Negative: | True Negative: |
| | **18** | **240** |

Accuracy = 88%    Precision = 80%    Recall = 87%
(358/404)          (118/146)          (118/136)

*Accuracy: Accuracy shows how often a classification ML model is correct overall.*[15]
*Precision: Precision shows how often an ML model is correct when predicting the target class.*[16]
*Recall: Recall shows whether an ML model can find all objects of the target class.*[17]

Our results yielded interesting data about the frequency of topics identified by the analysts versus the NLP topic modeling process. When comparing the categorization of documents, there were certain categories that the two approaches' evaluations were numerically aligned on (e.g., Movement of Forces/Materials where NLP identified 31 and analysts identified 30), examples where NLP identified far more of a category (Funding where NLP identified 80 and analysts identified 23), and where analysts identified far more than the NLP (Arms Control where NLP identified 65 and analysts identified 88).

Additionally, we wanted to consider how impacted data (documents with obfuscated text) would affect the NLP topic identification process. We reviewed each document to classify them as either clean or impacted, which left us with 351 clean and 53 impacted PDs. We then pulled results again only using clean PDs to compare how those results differed from our original data. The data comparison provided the following results:

| | Threat of Force | Movement | Maintenance | Arms Control | Monitoring/ Verification | Programs | Funding |
|---|---|---|---|---|---|---|---|
| **Analyst** | **3.96%** (16/404) | **7.42%** (30/404) | **7.42%** (30/404) | **21.78%** (88/404) | **12.62%** (51/404) | **15.84%** (64/404) | **5.69%** (23/404) |
| **NLP (All data)** | **5.69%** (23/404) | **7.67%** (31/404) | **17.08%** (69/404) | **15.84%** (65/404) | **8.66%** (36/404) | **30.94%** (125/404) | **19.80%** (80/404) |
| **NLP (Clean data)** | **5.98%** (21/351) | **8.26%** (29/351) | **15.38%** (54/351) | **14.25%** (50/351) | **7.41%** (26/351) | **29.63%** (104/351) | **19.09%** (67/351) |

While the data seemed to align in some categories where there were similar results from all three evaluations, there were instances where the NLP with both total and clean data identified close to four times as many documents in a category than analysts had (see Funding above). Our research team discussed these results and determined that the input we provided for the HITL aspect of the NLP programming, creating a tiered list of categories for each topic, may have influenced the program's topic identification patterns in unanticipated ways identifying topics differently than the analysts had. Specifically, we discussed how analysts did not have a set number of categories to identify per relevant document and were not limited to selecting three categories. This could have led to two outcomes: either NLP was limited to choosing three categories per document when the program could have identified the PD as relevant to even more categories or the NLP was forced to choose three categories even if less than three categories were represented in the document.

Additionally, the results we pulled were looking at the dataset as a whole and the topics contained within it to see how our judgments aligned. We did not do a document-by-document analysis to see how closely the NLP topic identification mirrored the analysts' judgments. A side-by-side comparison of analyst judgments and NLP results would provide insight into the in-depth reading of a document and would be a fruitful area of research for future study. Our goal was to judge if NLP could assist analysts in the data gathering phase, and regardless of the specific perceptions of the NLP on the contents of the documents if the program can identify them as relevant based on the analyst input the tool is useful to expedite the process.

## Discussion

The categories we chose to scan for (Threat of Force, Movement of Forces/Materials, Maintenance, Arms Control, Monitoring/Verification, Programs, and Funding) were all directly related to key aspects of strategic signaling, and our research showed that both analysts and NLP were able to identify language related to those topics in a significant number of PDs indicating that PDs were used to signal strategic intentions to the Soviet Union/Russia.

As we identified early in our data collection phase, we saw a sharp decline in the number of publicly available PDs per administration. As seen in the chart below, PDs from the Reagan administration where most available with 297, followed by Bush with 67, and Clinton with 40.

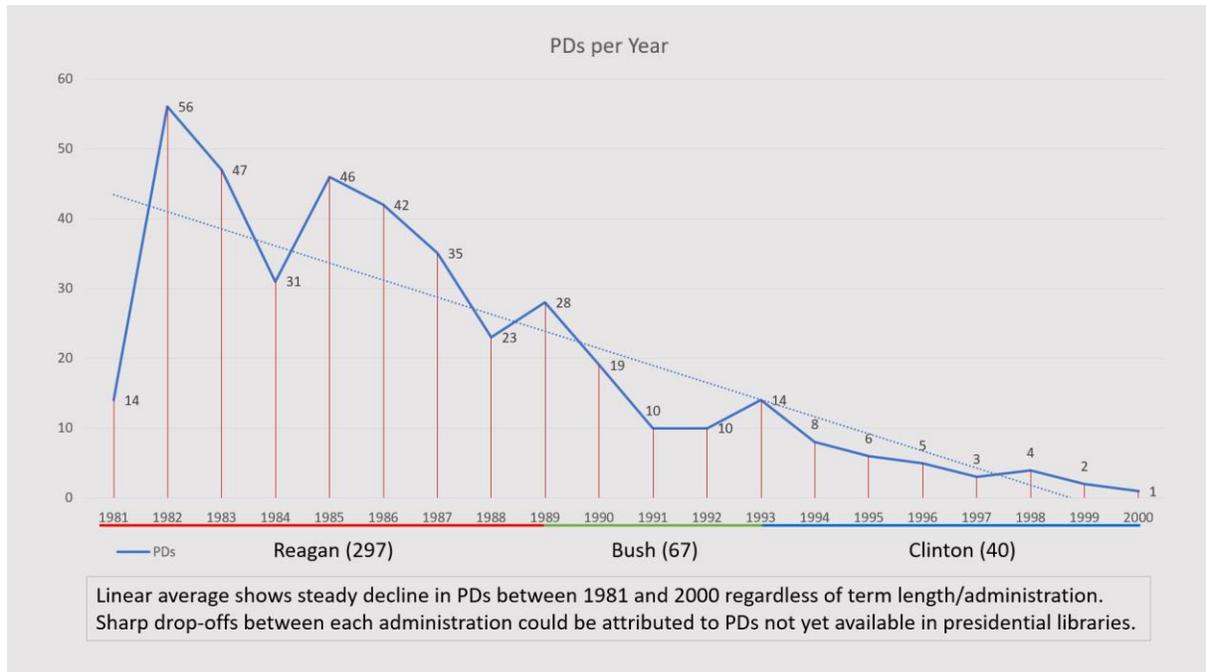

*Figure 1 [Chart shows available PDs per year divided by presidential administration.]*

While we had anticipated the data being difficult to obtain for more recent years, we did not account for how severe the data drop-off would be during the Clinton administration where over a seven-year period there were only 40 available PDs with only one available PD in 2000. This lack of data prevented us from comparing the composition of yearly patterns in signaling, as the year with the most PDs (1982- 56 PDs) and the least (2000- 1 PD) were incomparable. Despite this, an analysis of signaling patterns within an administration remained a feasible option.

The Reagan administration dataset had the highest number of PDs with 297 available documents. Within the available documents, analysts identified 111 (37.4%) relevant documents while NLP identified 115 (38.7%). Regarding analyst perceptions, reductions/arms control was the most frequently identified category with 75 documents marked as containing relevant information. This was followed by programs with 51 documents and monitoring/verification with 47 documents. Similarly, NLP also identified reductions/arms control as the most common category but returned a value of 101 relevant documents. However, NLP identified the second most common categories as "monitoring/verification" and "funding" with 56 documents each.

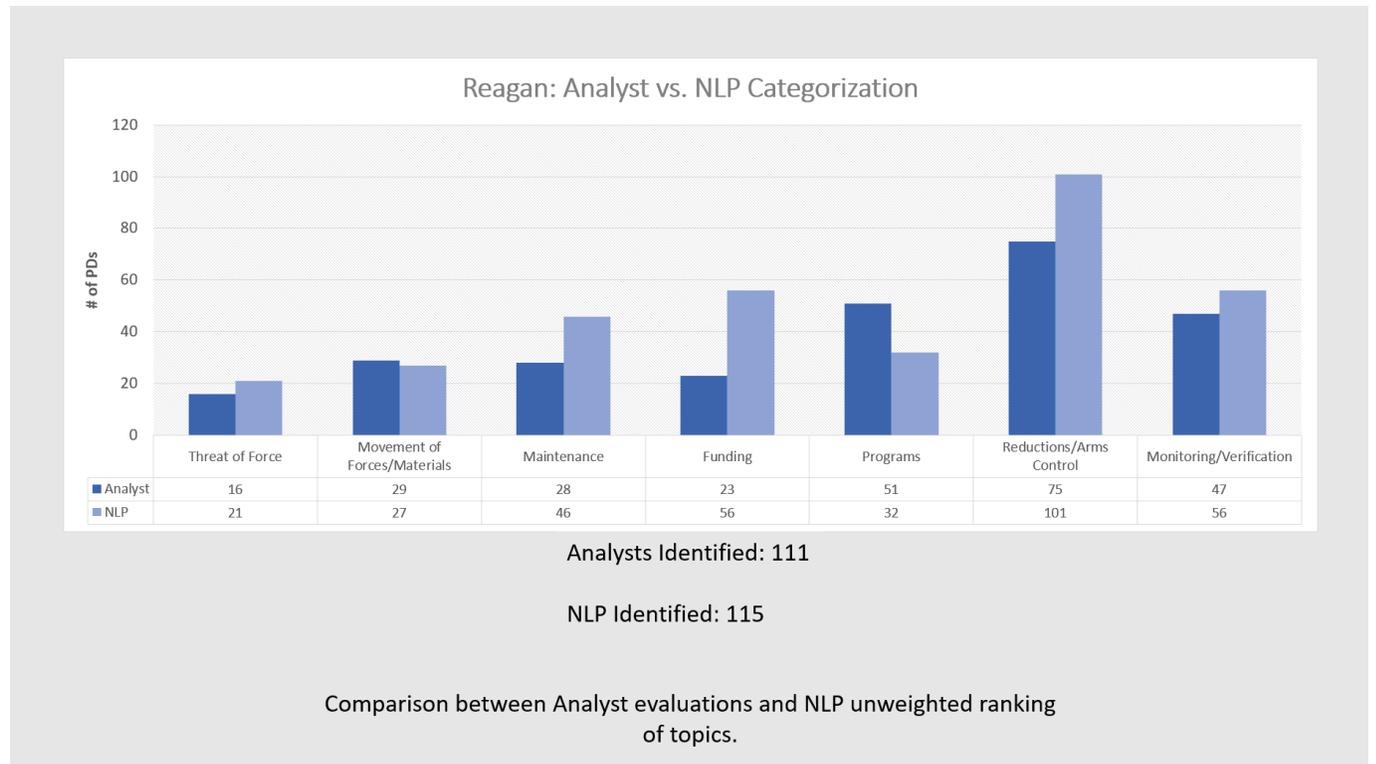

**Reagan: Analyst vs. NLP Categorization**

| | Threat of Force | Movement of Forces/Materials | Maintenance | Funding | Programs | Reductions/Arms Control | Monitoring/Verification |
|---|---|---|---|---|---|---|---|
| Analyst | 16 | 29 | 28 | 23 | 51 | 75 | 47 |
| NLP | 21 | 27 | 46 | 56 | 32 | 101 | 56 |

Analysts Identified: 111

NLP Identified: 115

Comparison between Analyst evaluations and NLP unweighted ranking of topics.

*Figure 2 [Chart shows a comparison of the PD categorization by analysts and NLP for the Reagan Administration.]*

The Bush administration dataset had 67 available documents, where analysts identified 17 (25.3%) relevant PDs and NLP identified 24 (35.8%). For analysts, the most common category was "programs" which was identified in 11 documents, followed by "reductions/arms control" (10) and "monitoring/verification" (3). For NLP, the most common categories were "reductions/arms control" and "monitoring/verification" which were both identified in 19 documents, followed by "maintenance" which was identified in 18 documents.

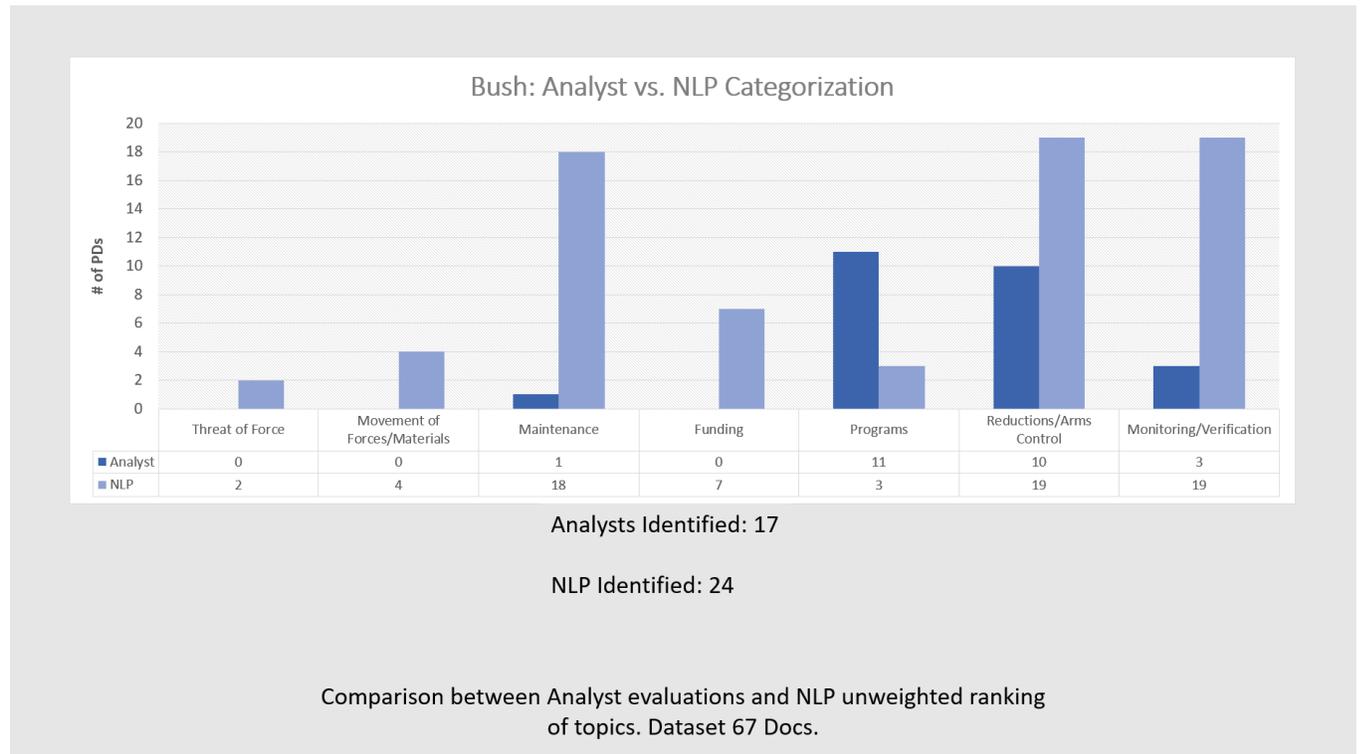

### Bush: Analyst vs. NLP Categorization

| | Threat of Force | Movement of Forces/Materials | Maintenance | Funding | Programs | Reductions/Arms Control | Monitoring/Verification |
|---|---|---|---|---|---|---|---|
| Analyst | 0 | 0 | 1 | 0 | 11 | 10 | 3 |
| NLP | 2 | 4 | 18 | 7 | 3 | 19 | 19 |

Analysts Identified: 17

NLP Identified: 24

Comparison between Analyst evaluations and NLP unweighted ranking of topics. Dataset 67 Docs.

*Figure 3 [Chart shows a comparison of the PD categorization by analysts and NLP for the Bush Administration.]*

The Clinton administration dataset had 40 available documents, where analysts identified five relevant PDs (12.5%) and NLP identified eight (20%). The most common category identified by analysts was "reductions/arms control" with three documents, followed by "programs" (2), and "maintenance" and "monitoring/verification," each with one. NLP identified "maintenance", "reductions/arms control", and "monitoring/verification" as the most common categories with five documents each. Interestingly, of the 40 documents available during the Clinton Administration, neither the analysts nor NLP identified any language on "threat of force" or "movement of forces/materials" in any of the PDs.

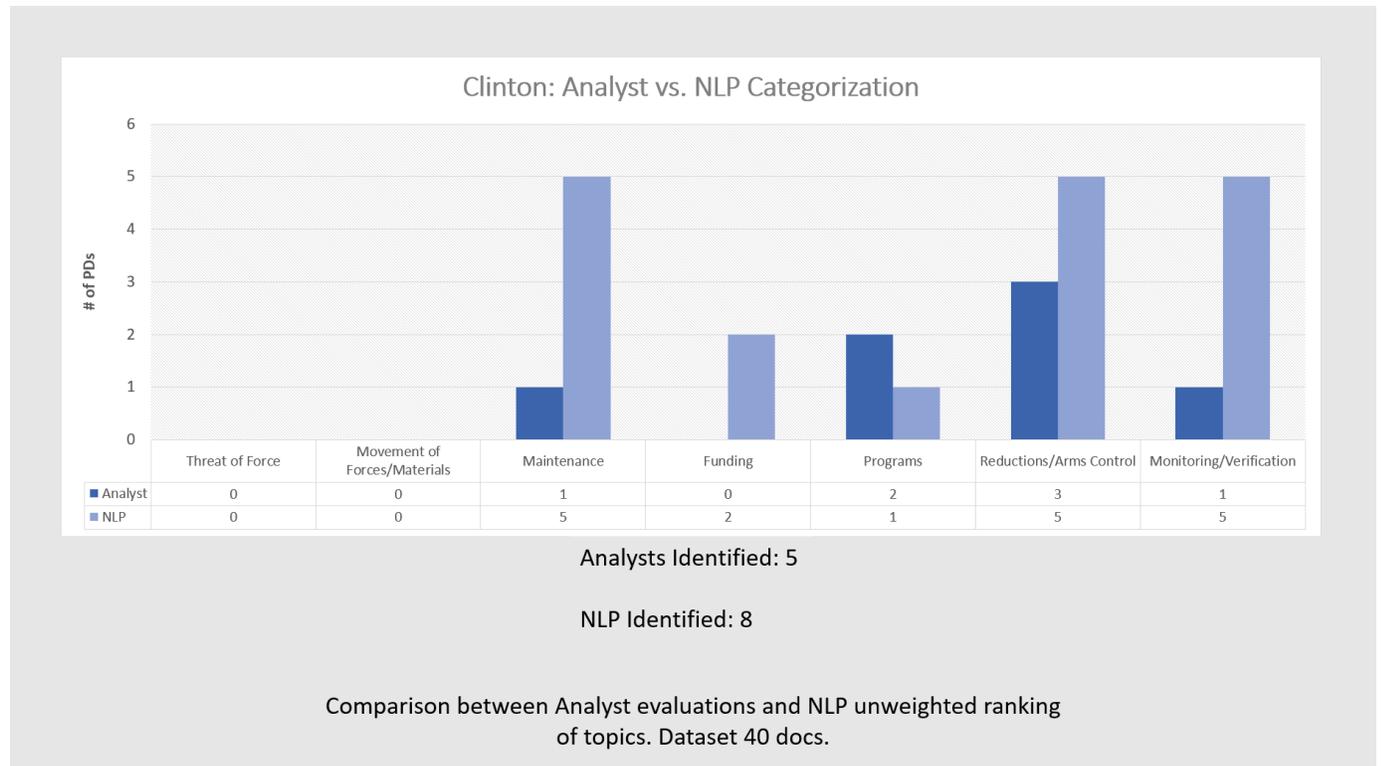

### Clinton: Analyst vs. NLP Categorization

| | Threat of Force | Movement of Forces/Materials | Maintenance | Funding | Programs | Reductions/Arms Control | Monitoring/Verification |
|---|---|---|---|---|---|---|---|
| Analyst | 0 | 0 | 1 | 0 | 2 | 3 | 1 |
| NLP | 0 | 0 | 5 | 2 | 1 | 5 | 5 |

Analysts Identified: 5

NLP Identified: 8

Comparison between Analyst evaluations and NLP unweighted ranking of topics. Dataset 40 docs.

*Figure 4 [Chart shows a comparison of the PD categorization by analysts and NLP for the Clinton Administration.]*

As we chose the categories based on our understanding of signaling and identified language without having previous knowledge of the contents of the PDs, based on the results showing the frequency of topics identified by both NLP and analysts it is clear that PDs served as signaling mechanisms where presidential administrations could use the documents to impact high-level, visible strategic policy.

The results of the case study indicated that using NLP to evaluate general topics within a corpus of written data could provide meaningful results in future applications, but we noted significant shortfalls in its application to our research. These shortfalls were mainly due to prompt engineering challenges as the NLP instructions were too broad in some aspects and too specific in others. There were a number of cases where NLP identified documents that were not relevant to the research, indicating we should have narrowed the instructions to find a more precise set of documents. There were also cases where NLP missed documents that were relevant, indicating the instructions were missing key elements that the analysts had considered relevant. Additionally, we only considered how topics could be identified within a corpus, but not how the topic identification lined up between analysts and NLP per document, so we did not determine if NLP and analysts were identifying the same topics in documents, a useful validation that could be considered in future work. However, based on the initial results of our research, we believe additional research into the application of this technology could provide a valuable tool for social science research.

In our specific use case, the application and evaluation of this tool was complicated by the lack of publicly available PDs in recent years. Because of this, we were hampered in drawing broader comparisons on signaling patterns in a comparative manner, limiting us to the task of presenting the frequency of topics identified, a more limited but still useful addition to the literature. As U.S. and Russian relations enter a period of tensions reminiscent of the Cold War and as the arms control infrastructure continues to deteriorate, studying documents like PDs could provide historical lessons on effective signaling measures for the U.S. to apply in the modern context. Without effective signaling, the boundaries set by the United States may not have the necessary weight to protect our interests and national security, which highlights the importance of work pursued in this paper.

## Conclusion

By manually reviewing the data and discrepancies, we determined that using NLP as the sole means to identify topics within a large dataset may result in missing data. However, by using NLP as a tool to evaluate general topics within a large set of written documents to understand general themes and target in-depth research, this technology does succeed in providing a bird's eye view of the general topics discussed before more time intensive research resources are dedicated to investigations.

One difference between analysts and NLP was that NLP would process all scannable text within a document while analysts would only read some sections of the documents in-depth if relevant concepts were identified. The purpose of the research was understanding how NLP can expedite topic identification, so analysts focused on quickly locating topics within each document but did not read every PD to the same degree that NLP processed available text, a key strength of NLP. However, analysts were able to read around document interference while NLP, even using OCR, did not have a clean scan of the data.

Further refinement of the NLP programming could result in more accurate results, but we still consider the present work successful as our programming was designed to test the ability of the technology to pull relevant topics from documents, a task this model succeeded at, not to perfectly mirror the analysts' ability to read and interpret the information, a topic for further work. Additional research in this field could provide insight into alternative methodologies which would reduce time or costs to conduct similar research and may provide different results in NLP feedback.

## 2025 Epilogue on Evolving AIML Tools

This research was conducted between 2023 and 2024, and the results indicate that the social sciences could potentially leverage tools similar to—or better than—the ones we used to significantly expedite the research process. Analysts spent a significant amount of time reading through documents, many of which turned out to be irrelevant to the scope of our project. This time could be cut down by leveraging AI tools like Semantic Search, which understands the meanings and nuance of words rather than simply searching for text matching, or other similar capabilities.[18] AI-enabled social science research could cut down the time required to collect sources for literature reviews and expedite the background research process, saving time and money on projects and improving productivity. Our project shows the applications and limits of one specific technology, but future applications of our framework using improved methods may yield entirely different results.

---

[18] *What is Semantic Search?: A comprehensive semantic search guide*. Elastic. (n.d.). https://www.elastic.co/what-is/semantic-search. Accessed 23 October 2025.

# Annex 1: Analyst Instructions for Document Review

ANALYST INSTRUCTIONS

1. If the first section of the document is related to the nuclear enterprise (including DOD) or related directly to arms control, then "contains_nuclear=1". If document is not related to above conditions, then "contains_nuclear=0"
   a. Read first section of document to gather information on general topic, search for "nuclear" using document search function
2. Repeat process on PDs of other teammates (divided into thirds) to challenge/confirm evaluation of documents using Step 1. Write a brief summary of why the document is relevant in a separate column.
3. Note Potential Categorization within the relevant column.
   a. Category breakdown:
      i. Threat of Force
         1. Language ensuring reliability of damage
         2. Guarantee efficiency
         3. Aggressive language
      ii. Movement of Forces/Materials
         1. Movement/relocation of completed forces
         2. Movement of material for development or testing
      iii. Maintenance
         1. Includes redesign, testing, components
      iv. Funding
         1. Budgetary allocations for new programs
         2. Investment into existing programs
      v. Programs
         1. Authority approval for new programs (i.e. FY___ docs)
         2. Status updates on existing programs
      vi. Reductions/Arms Control
         1. Relates specifically to goals, negotiations, outcomes, or all related processes in arms control/reduction
      vii. Monitoring/Verification
         1. Explicitly says the words verification/monitoring
4. Read PDs of other teammates (final third of the documents) to evaluate results of all previous steps, provide input on uncertainties. Provide a secondary summary of why the document is relevant in a separate column. Expand judgment of categories to all relevant categories in the document and mark them in the form with an "X" as necessary.

5. Divide PDs by year, administration, and category to perform analysis

6. Determine total values of relevant outputs to compare to NLP outputs

# Annex 2: Complete Research Plan

RESEARCH PLAN/ORDER OF RESEARCH PROCESSES

1. Analysts create a shared Excel file and compile all available PDs[19]
2. Analysts collectively decide what we're looking for as signaling
   a. "We will be looking for documents related to the nuclear enterprise (including DOD) or related directly to arms control, henceforth strategic signaling"
3. Analysts coordinate and select a list of categories we determined to be essential for strategic signaling
   a. Category breakdown:
      i. Threat of Force
         1. Language ensuring reliability of damage
         2. Guarantee efficiency
         3. Aggressive language
      ii. Movement of Forces/Materials
         1. Movement/relocation of completed forces
         2. Movement of material for development or testing
      iii. Maintenance
         1. Includes redesign, testing, components
      iv. Funding
         1. Budgetary allocations for new programs
         2. Investment into existing programs
      v. Programs
         1. Authority approval for new programs (i.e. FY___ docs)
         2. Status updates on existing programs
      vi. Reductions/Arms Control
         1. Relates specifically to goals, negotiations, outcomes, or all related processes in arms control/reduction
      vii. Monitoring/Verification
         1. Explicitly says the words verification/monitoring
4. Assign a numbering system across all PDs for uniformity in data analysis
5. Compile all PDs as PDFs into a singular folder location
6. Convert all PDs to text
7. The following steps occur simultaneously
   a. NLP Application
      i. Use the Latent Dirichlet Allocation (LDA) to extract the topics hidden within the corpus

---

[19] Exclusively using presidential libraries

  ii. Assign topics to a relevant category based on identified language (HITL phase)
    1. Topic language categorization should align with language used for Analyst categorization
    2. Categorization is done in a ranked manner with a primary, secondary, and tertiary category per topic
    3. If the topic is not applicable, it is categorized under "Other"
  iii. Identify the following in data outputs (See Results Excel)
    1. Frequency of topics within PDs
    2. Binary selection of top three categories per PD
    3. Weighted frequency of categories per PD
    4. Whether the document contains the word nuclear (contains_nuclear=0/1)
    5. Whether the "Other" category dominates the defined categories (If WC_7 >/= .50 then other_dominates=1)
    6. If the program believes the document should be read by analysts (contains_nuclear=1, other_dominates=0)
  iv. Determine total values of relevant outputs to compare to analyst outputs
 b. Analyst instructions
  i. If the first section of the document is related to the nuclear enterprise (including DOD) or related directly to arms control then "contains_nuclear=1". If document is not related to above conditions, then "contains_nuclear=0"
    1. Read first section of document to gather information on general topic, search for "nuclear" using document search function
  ii. Repeat process on PDs of other teammates (divided into thirds) to challenge/confirm evaluation of documents using Step 1. Write a brief summary of why the document is relevant in a separate column.
  iii. Note Potential Categorization within the relevant column.
  iv. Read PDs of other teammates (final third of the documents) to evaluate results of all previous steps, provide input on uncertainties. Provide a secondary summary of why the document is relevant in a separate column. Expand judgment of categories to all relevant categories in the document and mark them in the form with an "X" as necessary.
  v. Divide PDs by year, administration, and category to perform analysis
  vi. Determine total values of relevant outputs to compare to NLP outputs
8. Identify discrepancies in data, review all cases for analyst/NLP and mark documents accordingly
9. Compare data outputs by category and administration.